\documentclass{article}
\usepackage{spconf,amsmath,graphicx}

\title{SMALL NOISY and PERSPECTIVE FACE DETECTION USING DEFORMABLE SYMMETRIC GABOR WAVELET NETWORK}
\name{Sherzod Salokhiddinov and Seungkyu Lee}
\address{Dept. of Computer Engineering, Kyung Hee University, Republic of Korea}
\usepackage{amsrefs}
\newenvironment{rezabib}
  {\bibdiv\biblist\setupbib}
  {\endbiblist\endbibdiv}

\def\setupbib{\catcode`@=\active}
\begingroup\lccode`~=`@
  \lowercase{\endgroup\def~}#1#{\gatherkey{#1}}
\def\gatherkey#1#2{\gatherkeyaux{#1}#2\gatherkeyaux}
\def\gatherkeyaux#1#2,#3\gatherkeyaux{\bib{#2}{#1}{#3}}

\begin{document}
%
\maketitle
\begin{abstract}
Face detection and tracking in low resolution image is not a trivial task due to the limitation in the appearance features for face characterization. Moreover, facial expression gives additional distortion on this small and noisy face. In this paper, we propose deformable symmetric Gabor wavelet network face model for face detection in low resolution image. Our model optimizes the rotation, translation, dilation, perspective and partial deformation amount of the face model with symmetry constraints. Symmetry constraints help our model to be more robust to noise and distortion.
Experimental results on our low resolution face image dataset and videos show promising face detection and tracking results under various challenging conditions.
\end{abstract}
\begin{keywords}
Face Model, Perspective, Face Detection, Gabor Wavelet, Tracking
\end{keywords}
\section{Introduction}
\label{sec:intro}
Face detection and tracking have been essential topic in computer vision and image processing.
In order to detect human face robustly, abstracted human face models are built and detected in images or videos by minimizing distance between the model and face region. Viola et al. \cite{viola2004robust} propose real time face detection algorithm using integral image and learning method with AdaBoost. However, this works well with only frontal faces.
Zhu et al. \cite{zhu2012face} propose a unified model for face detection using mixture of trees with a shared pool of parts.

Several researchers have proposed wavelet based model for face detection and tracking ~\cite{kruger2000efficient} ~\cite{feris2000tracking}.
Volker Kruger et al. \cite{kruger2000efficient} introduce Gabor wavelet network as an abstracted face model for fast and efficient face tracking. This modeling finds frontal face, but it has difficulties in detecting faces with any perspective. They improve the method for affine transformed face detection \cite{krueger2000affine}.
Park and Lee \cite{park2008face} adapt machine learning method for accurate face detection.
Recently, Lee \cite{lee2011frequency} has proposed a facial symmetry based Gabor wavelet model for robust face detection which inspired our proposed method.
\begin{figure}[t]
\begin{minipage}[b]{1.0\linewidth}
  \centering
  \centerline{\includegraphics[width=8.5cm]{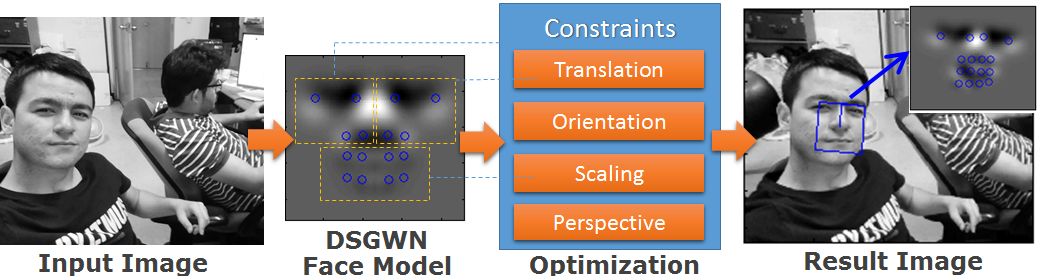}}
 \end{minipage}
\caption{Sample perspective face detection result with deformable symmetric GWN model and constraints}
\label{fig:pers}
\end{figure}

Detecting face from low resolution and noisy images is challenging task, because they contain very little appearance clue for face detection.
In order to overcome this problem, Qahwaji et al. \cite{qahwajia2001detecting} obtain boundary information of a target in an image and then attempt to eliminate noise using Hit-Miss transform. Note that the test images have uniform background.
Chung et al. \cite{hsu2007improved} concentrate on face detection in low resolution videos.
Face detection based on the modeling using the mixture of tree and a shared pool of parts has been proposed by Zhu et al.\cite{zhu2012face} to address the face localization problem occurring in the images with different viewpoints.
Yang et al. \cite{yang2013occluded} studied face detection in low resolution and occluded images. They propose three-layer hierarchical model which consists of six nodes (head and shoulder, face, left eye, right eye, nose, mouth).
Zheng et al. \cite{zheng2010face} present 12-bit modified census transform for face detection in low resolution color images.
Kruppa and Schiele \cite{kruppa2003using} use local context which is trained with instances that contain an entire head for face detection in low resolution images.
Mahdi et al. \cite{rezaei2014global} use image denoising method which is suggested by Kovesi \cite{kovesi1999phase}.

In this paper, we propose deformable symmetric Gabor wavelet network (DSGWN) model for face detection in low resolution and noisy images. Advantages of the GWN face model that has been used in many previous work \cite{lee2011frequency} is that we are able to build a scalable face model with controlled number of wavelets for coarse to fine representation of human face.
In real world images, faces almost always contain perspective distortion. Particularity, people take a picture of human face close to them frequently showing more serious distortions. With small sized image, perspective distortion and others like facial expression seriously degrade the face detection performance. Furthermore, our symmetric constrains keep the global facial shape without detailed appearance features in small image.
In our modeling, we build symmetric GWN model including perspective view parameter. Then we separate wavelets to several groups to allow deformation in face structure. We add symmetry constraints between these groups. In the detection and tracking, we adapt the frequency guidance detection method proposed in \cite{lee2011frequency}.

\section{PROPOSED METHOD}
\label{sec:format}
\begin{figure}
  \centering
  \centerline{\includegraphics[width=8.5cm]{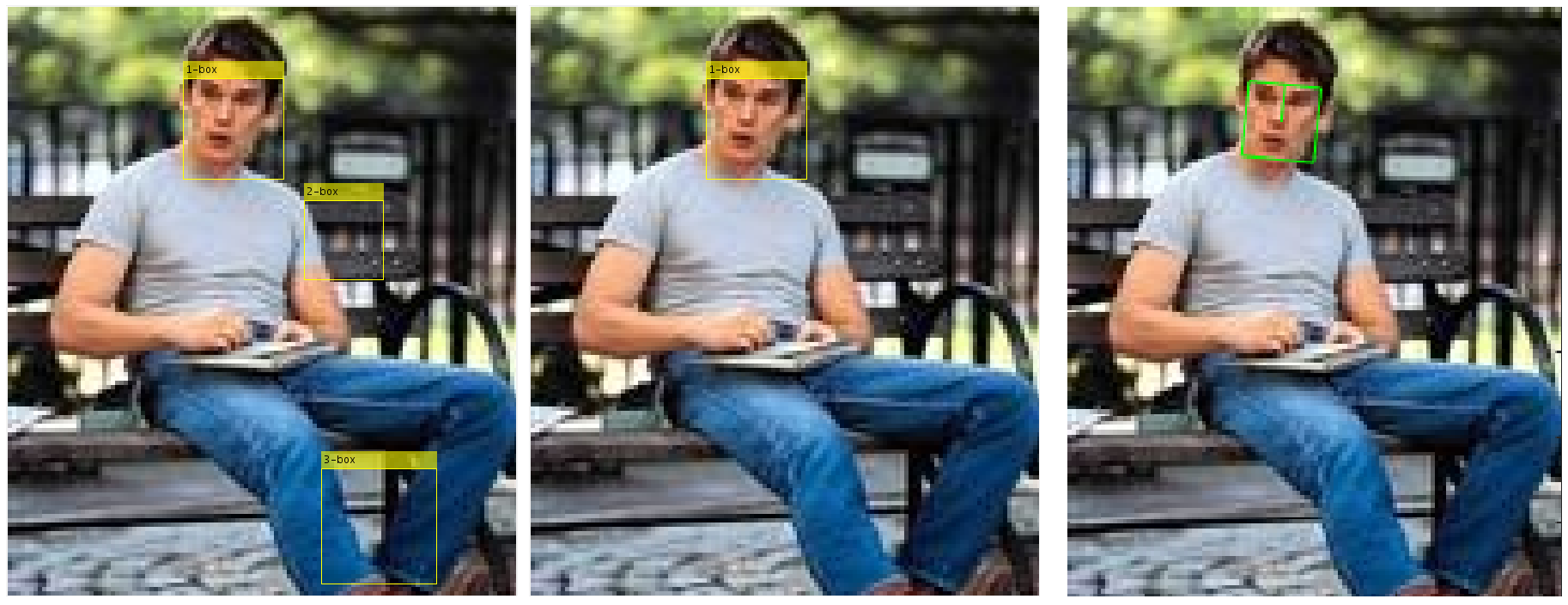}}
  (a) candidate windows \hspace{0.2in} (b) CNN classifier \cite{lee2011frequency} \hspace{0.2in} (c) our DSGWN
\caption{Stages of proposed method}
\label{fig:methods}
\end{figure}

Proposed method consist of three stages as shown in Figure \ref{fig:methods}, i.e. searching face-like regions, face classification and find accurate face information(accurate location, rotation, scaling, perspective, symmetric etc). In first stage we find face-like regions. In this section we use object detection method which is Viola-Jones object detection framework \cite{viola2001rapid} \cite{viola2001robust}. By minimizing detection rate we can get more face-like regions. It helps to find face regions from low resolution images.
In the second stage, we use Convolution Neural Network classification. Because we get several candidate face regions from first stage. Then we should keep only face regions. 

In the last stage using GWN face model we find accurate shape and position of face. In this stage we generalize symmetric GWN face model \cite{lee2011frequency} by incorporating perspective distortion and extending it to be our deformable symmetric model (DSGWN) with geometric and symmetric constraints.
In order to recognize perspective distortion, our optimization has extended searching space including parameters of perspective transformation. Our GWN model has three subgroups (top-left, top-right and bottom) with perspective, translation, orientation and size parameters. Each subgroup try to find optimized location searching for optimal appearance of human face under constraints such as distance, orientation difference and size difference between subgroups keeping symmetric structure of our model.

\subsection{FINDING CANDIDATE FACES}
\label{sec:find}
In order to get roughly face regions from low resolution images we use Viola-Jones object detection method \cite{viola2001robust}.  This method is robust very high detection rate (true-positive rate)  very low false-positive rate always.  Basic idea of this algorithm is sliding window across image and evaluate a face model at every location.
This algorithm has four stages:

    1. Haar feature selection

    2. Creating an Integral Image

    3. Adaboost Training

    4. Cascading Classifiers

Concept of integral image helps to calculate Haar features quickly. 

In order to select features we use Adaboost algorithm. Boosting is a classification scheme that works by combining weak learners into a more accurate ensemble classifier.  We can define weak learners based on rectangle features as follows:

\begin{equation}
h_{t}(x)= \left\{\begin{matrix}
1 \ if p_{t} f_{t} (x)> p_{t}  \theta _{t} ) \\
0 \ otherwise 
\end{matrix}\right.
 \end{equation}

here $x$ is window, $h_{t}$ is value of rectangle feature, $p_{t}$ is parity and   $\theta _{t}$ is   threshold. Adaboost learning algorithm creates a small set of only the best features to create more efficient classifiers \cite{freund1997decision}.

Then in order to achieve increased detection performance and reduce computation time we use cascade classifiers. A cascade is a way of combining classifiers in a way that a given classifier is only processed after all other classifiers coming before it have already been processed. In last stage threshold value may  be adjust to tune accuracy. Choosing lower threshold helps to find face regions from low resolution images, but more probably we get false positive windows as shown in Figure \ref{fig:methods} (a).

\subsection{FACE CLASSIFICATION}
\label{sec:subhead1}
From first stage we get candidate windows consist of positive and negative. Next thing to do is get only face-like regions. In order to check candidate windows we use CNN classification. In order to make structure of CNN we use LeNet-5  \cite{lecun1998gradient}. Our CNN network is consist of 8 layers, 3 convolution layers, 3 max pooling layer and 2 fully connected layers (shown as figure 3-not yet). Output is two which are face and non-face. Input is 32x32 gray-scale image, because we need find small faces that's why input is small. For training of CNN face classification we use 1000 positive and 4000 negative images. In this stage, the positive proposed candidate windows are selected by CNN face classification (as shown in Figure \ref{fig:methods} (b) .

\subsection{GWN FACE MODEL}
\label{sec:subhead2}
Gabor wavelet network (GWN) used by Kruger et al. \cite{krueger2000affine} for face detection optimize the location, orientation and size of each Gabor wavelet based on the local appearance of human face. In order to build the GWN for frontal face, we use \textit{N} wavelets $\Psi=$\big\{$\psi_{\textbf{n}  1}$, $\psi_{\textbf{n} 2}$, ... ,$\psi_{\textbf{n} N}$\big\} with parameter vectors $\textbf{n}= \left ( c_{x},c_{y}, \theta ,s_{x},s_{y}  \right ) ^{T}$ calculated as follows \cite{lee2011frequency}. Structure of network shown as figure [make], which consist of one input layer  
\[
\varphi_{n}(x,y)=exp(-\frac{1}{2}(\left [ s_{x}((x-c_{x}) \cos \theta -(y-c_{y}) \sin \theta  \right )]^{2}
\]
\[
+\left [ s_{y}((x-c_{x}) \sin \theta + (y-c_{y}) \cos \theta ) \right ]^{2})
\]
\begin{equation}
\times \sin(s_{x}((x-c_{x}) \cos \theta -(y-c_{y}) \sin \theta ))
\label{eq:wavelets}
\end{equation}
where $(c_{x},c_{y})$ is translation, $\theta$ is orientation and $(s_{x},s_{y})$ is wavelet size.
Based on a given frontal face image $I$, GWN is built by minimizing the energy function $E$ with respect to weight $\omega _{i}$ and $n_{i}$ \cite{lee2011frequency}.
\begin{equation}
E= \underset{n_{i}\ \omega _{i}\ for\ all \ i}{min} \left \| I-\sum_{i}\omega _{i}\varphi_{n_{i}} \right \|_{2}^{2}
\label{eq:energy_f}
\end{equation}

Equation (\ref{eq:energy_f}) shows that the $\omega_{i}$ and the wavelet parameter vector $n_{i}$  are optimized. Two vectors which
$\Psi=\big\{\psi_{\textbf{n} 1}, \psi_{ \textbf{n}2}, ... ,\psi_{ \textbf{n} N}\big\} ^{T}$
and $w=\left ( \omega_{1},\omega_{2},...,\omega_{N} \right )^{T}$ define the Gabor wavelet network  $\left ( \varphi ,w \right )$ for image $I$.

\subsection{PERSPECTIVE GWN FACE MODEL}
\label{sec:subhead}
Homography describes the transformation function between two images having respective perspective distortions that has been used for panorama creation stitching multiple images taken at different positions. This transformation is constructed in homogeneous coordinates represented by a 3x3 matrix with 8 degrees of freedom (DoF). We represent projective transformation of face in an image using homography matrix.
The advantage of utilizing homogeneous coordinates is that we can represent position of each wavelet as a tuple $(x,y,w)$, where $w$ is a scale parameter. For simplification, we will let $w$ be fixed at 1. Homogeneous coordinates allow us to perform an image projective transformation by using standard matrix multiplication.

\begin{equation}
\begin{bmatrix}
x^{'}\\
y^{'}\\
w^{'}\\
\end{bmatrix}=
\begin{bmatrix}
 h_{11}&h_{12}&h_{13} \\
 h_{21}&h_{22}&h_{23} \\
 h_{31}&h_{32}&h_{33} \\
\end{bmatrix}
\begin{bmatrix}
x\\
y\\
1\\
\end{bmatrix}
\label{eq:homography}
\end{equation}
where $x$ and $y$ are current locations of Gabor wavelet, $x^{'}$ and $y^{'}$ are projected locations of Gabor wavelet, and $w^{'}$ is scale parameter.
All projected points have been calculated by using equation (\ref{eq:homography}). In order to recover original coordinates, each elements of tuple are divided by its homogeneous scale parameter. Then as we mentioned above for the simplification we set $w$ to 1 (Equation (\ref{eq:point}))
\begin{equation}
\frac{ \left \langle x^{'},y^{'},w^{'} \right \rangle}{w^{'}}= \left \langle \frac{x^{'}}{w^{'}},\frac{y^{'}}{w^{'}},\frac{w^{'}}{w^{'}} \right \rangle=\left \langle \frac{x^{'}}{w^{'}},\frac{y^{'}}{w^{'}},1\right \rangle=\left \langle x^{''},y^{''} \right \rangle
\label{eq:point}
\end{equation}
where, $x^{''}$ and $y^{''}$ are normalized projected location of Gabor wavelets.
By choosing the correct values for the homography matrix, we could get a transformed GWN face model (A sample is overlapped on the result image in Figure \ref{fig:pers}).

\subsection{DEFORMABLE GWN FACE MODEL}
\label{elastic:subhead}
\begin{figure}[b]
  \centering
  \centerline{\includegraphics[width=8.5cm]{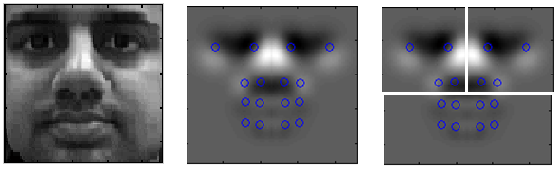}}
  (a) Frontal face \hspace{0.2in} (b) BSGWN \cite{lee2011frequency} \hspace{0.2in} (c) Our DSGWN
\caption{Our DSGWN model compared with BSGWN \cite{lee2011frequency}}
\label{fig:DGWN}
\end{figure}
Now we have symmetric GWN face model with perspective parameter.
In order to allow deformation in the structure of the face, we separate Gabor wavelets to three group (A.top-left, B.top-right and C.bottom) as shown in Figure \ref{fig:DGWN} (c). Each group has parameter vector $m=(c_{x},c_{y},\theta ,s_{x},s_{y})$ that consists of translation, orientation and dilation. In this work, we use 16 Gabor wavelets. Group A and B consist of 4 wavelets and group C consists of 8 wavelets. At each group, respective group translation, orientation and dilation values are calculated. In our face detection and tracking, each group separately finds best fit in local area keeping relations defined and controlled by the symmetric constraints that will be explained in detail in the following section.
\section{DSGWN FACE DETECTION and Tracking}
\label{sec:pagestyle}
Deformable Symmetric GWN (DSGWN) face model is constructed with a perspective distortion. Then we calculate constraints for face detection with DSGWN. Our constraints consist of translation, rotation and scale differences between groups.
\begin{figure}[t]
  \centering
  \centerline{\includegraphics[width=8.5cm]{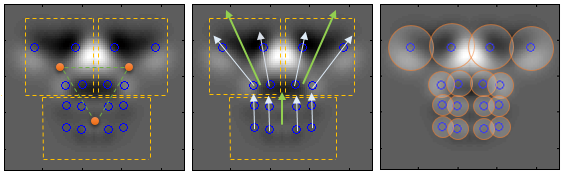}}
  a) Distance \hspace{0.3in} (b) Orientation \hspace{0.3in} (c) Scaling
\caption{Subgroup constraints of our DSGWN face model}
\label{fig:constraints}
\end{figure}
In order to calculate distance between groups we calculate average location of wavelets of each group $x_{c}=\frac{x_{1}+x_{2}+..+x_{i}}{i}$, $y_{c}=\frac{y_{1}+y_{2}+..+y_{i}}{i}$.
And we calculate Euclidean distance $d=\sqrt{(x_{2}-x_{1})^{2}+(y_{2}-y_{1})^{2}}$ between these average locations.
And then we calculate orientation difference between the subgroups. In order to calculate the orientation of each group A and B, we calculate upward orientations using 1st-3rd and 2nd-4th wavelet pairs ($\theta=atan2\left ( y_{2}-y_{1}, \ x_{2}-x_{1} \right )$) and average them as shown in figure \ref{fig:constraints} (b). For group C, we calculate orientations using 1st-5th, 2nd-6th, 3rd-7th and 4th-8th wavelet pairs. Orientation difference between groups $\theta _{d}=\left | \theta_{n1}-\theta_{n2} \right |$ are calculated using the group orientations.

Next constraint is scaling. Let's $\gamma_{h}$ and $\gamma_{v}$ are horizontal and vertical scale factors of an wavelet in our model. We have original scale of each Gabor wavelet. We calculate sum of multiplied scale of wavelets in each group as shown follows.
\begin{equation}
s_{n}=\sum_{i}\gamma_{h}(i)\gamma_{v}(i)
 \label{eq:scale}
\end{equation}
where $i$ is the index of wavelet for $n_{th}$ group. Then we get the scale difference between each groups $s=\left | s_{n1}-s_{n2} \right |$. In our face detection and tracking, we put respective constraints based on these relations. It helps to keep structure of symmetric face model while we allow deformation of our face model.

For our detection step, we build two types of super-wavelet: with all wavelets $\Psi_{n}^{all}$ and with the wavelets of each group $\Psi_{n}^{group}$ as a linear combination of wavelets.
\begin{equation}
\Psi_{n}^{all}(x)=\sum_i\omega_{i}\psi_{i}(sHR(x-\mathbf{c}))
 \label{eq:super}
\end{equation}
\begin{equation}
\Psi_{n}^{group}(x)=\sum_i\omega_{i}\psi_{i}(sR(x-\mathbf{c}))
 \label{eq:super-w}
\end{equation}
where $H$ is homography matrix, $s$ is scaling factor, $R$ is rotation matrix and $\mathbf{c}$ is translation vector. We use equation (\ref{eq:super}) for all wavelets in our model and (\ref{eq:super-w}) for each group`s wavelets.
Now, we optimize weight differences between the super-wavelets (face model) and an image or a current video frame. For our optimization, we use Levenberg-Marquardt algorithm.
\begin{figure}[t]
  \centering
  \centerline{\includegraphics[width=8.5cm]{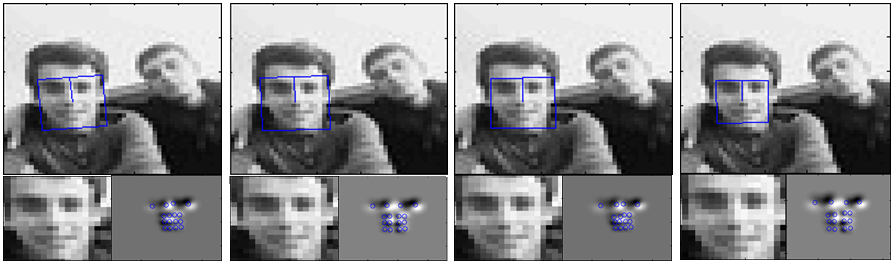}}
  \centerline{\includegraphics[width=8.5cm]{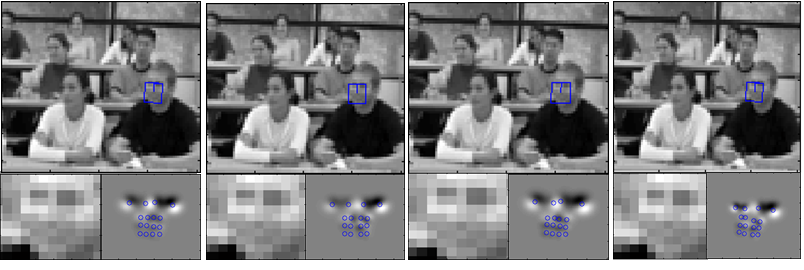}}
  \centerline{\includegraphics[width=8.5cm]{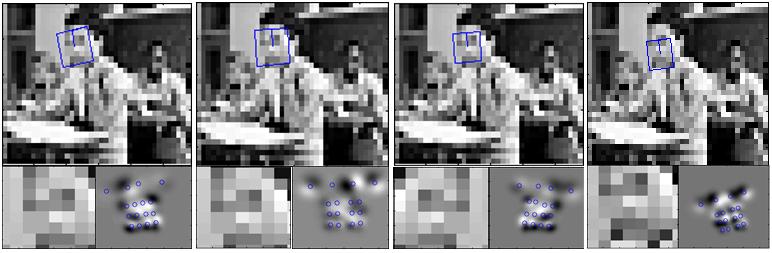}}
  (a) GWN ~\cite{kruger2000efficient} \hspace{0.1in} (b)BSGWN ~\cite{lee2011frequency} \hspace{0.1in} (c) DGWN \hspace{0.1in} (d) DSGWN
\caption{Face detection results compared with two previous face models: GWN ~\cite{kruger2000efficient} and BSGWN ~\cite{lee2011frequency}}
\label{fig:comparison}
\end{figure}
\section{Experimental Results}
Our DSGWN face model for face detection consists of 16 wavelets.
In figure \ref{fig:comparison}, four different results are compared. First column is the result of [3] and the second column is the result of [7]. In the third column, deformable face model is used without symmetric constraints and fourth column is the result of our proposed method with DSGWN. Detected face region(bottom-left) and transformed face model(bottom-right) show that all methods works properly with frontal face with the first image. On the other hand, the proposed method captures perspective and rotation of target human face even with very low resolution images compared to previous methods (second and third images).
\begin{figure*}[t]
\includegraphics[width=1\linewidth]{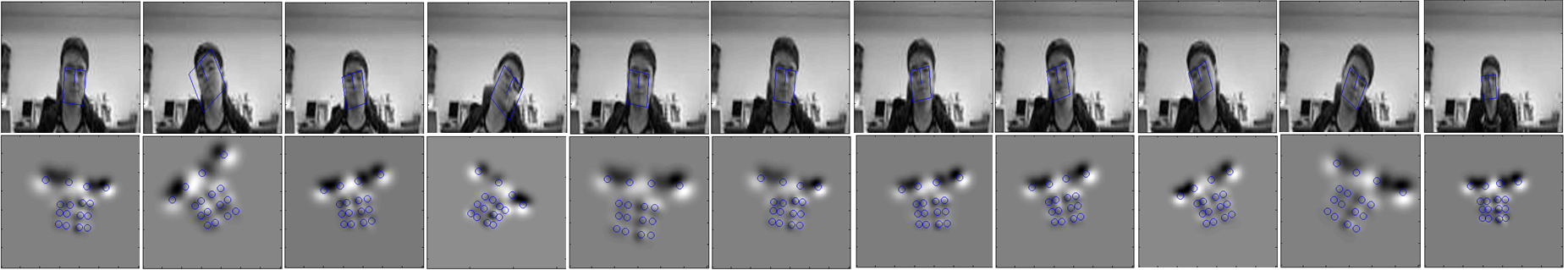}\\
\includegraphics[width=1\linewidth]{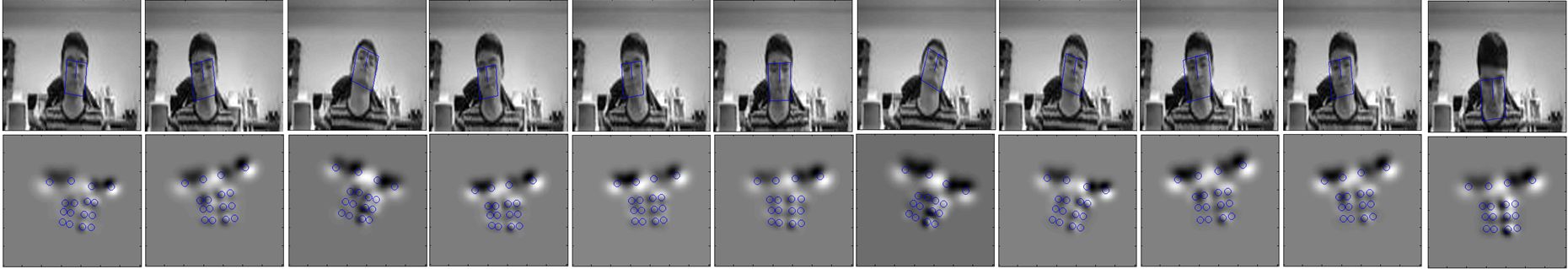}
   \caption{Tracking result with low resolution video showing significant rotation and perspective distortions}
   \label{fig:videores}
\end{figure*}
\begin{figure}[!hb]
   \begin{center}
   \includegraphics[width=1\linewidth]{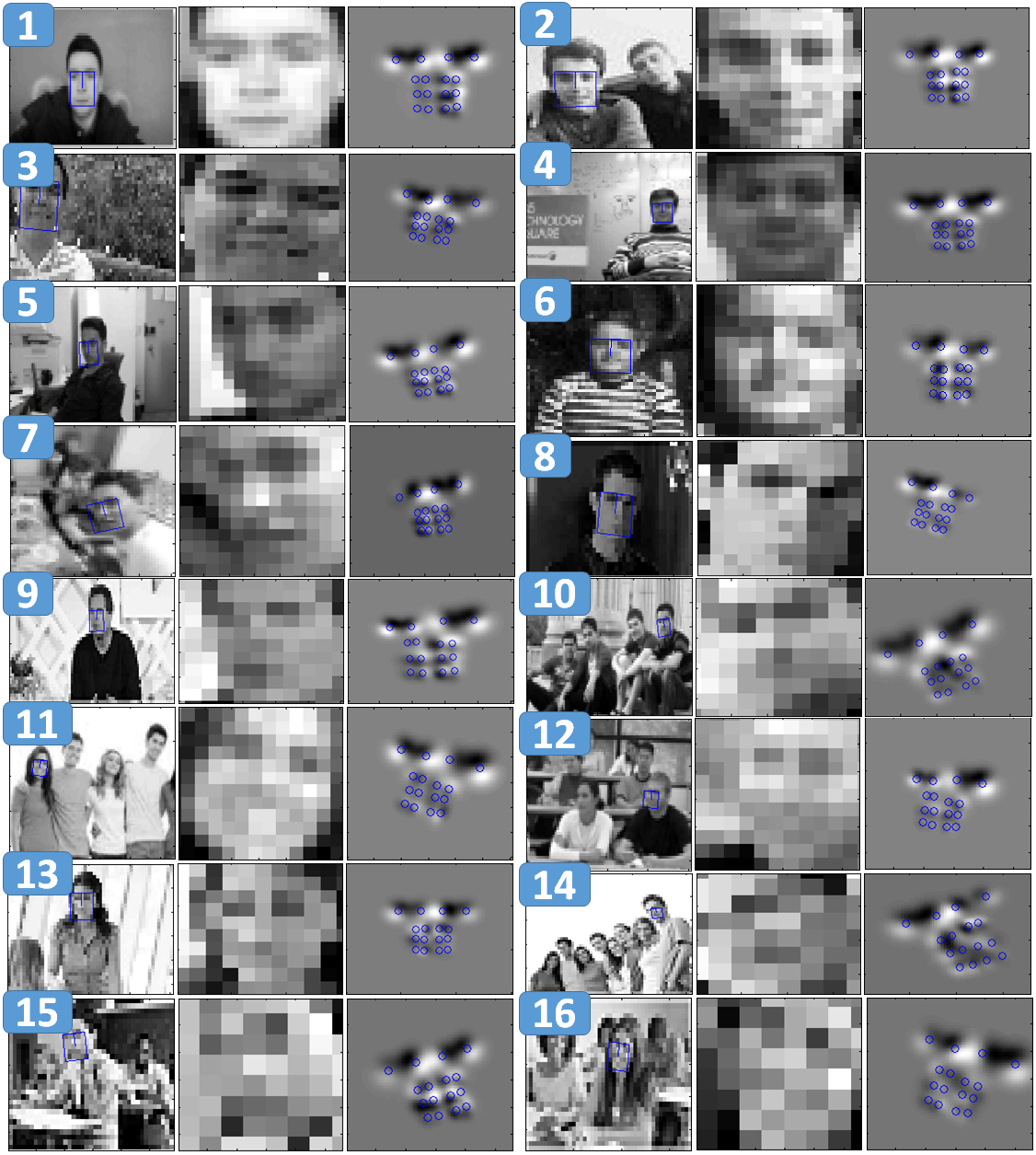}
   \end{center}
   \caption{Face detection results of our DSGWN face model with various still images}
   \label{fig:stillres}
\end{figure}
Figure \ref{fig:stillres} shows face detection results with various still images. Each result compares input image with detected face (blue square), detected face region image and transformed our DSGWN face model.
Note that we sorted the results by the resolution of detected face (fine to coarse).
Rotation (\#3, \#8, \#11, \#14, and \#16) and perspective (\#5, \#9, \#10, \#12, and \#16) faces are well detected and represented by our face model. Images \#14, \#15, and \#16 contains very low resolution faces and even it is difficult to recognize any detail of human face in the detected face images. Proposed method, however, finds global symmetry face shape showing reasonably correct detection results that can be checked in the input images. Especially, in the result of \#16, we observe that the perspective of human face is correctly detected with such small sized face image. Figure \ref{fig:videores} shows several screenshots of the tracking result from videos containing various rotation and perspective motions.
Video frames with perspective face (4th, 9th of 1st row and 3rd, 4th, 7th, 8th, 9th, 10th, 11th of 2nd row) faces are well tracked through the video.

\section{CONCLUSION}
\label{sec:foot}
In this paper, we propose DSGWN face model for face detection in low resolution and nosy images. Our model optimizes the rotation, translation, dilation, perspective and partial deformation amount of the face model with symmetry constraints. Our experimental results show that our method gives promising detection results compared with previous methods \cite{kruger2000efficient} \cite{lee2011frequency} in low resolution images and videos.


\nocite{*}

\begin{rezabib}
 
@article{viola2004robust,
  title={Robust real-time face detection},
  author={Viola, Paul and Jones, Michael J},
  journal={International journal of computer vision},
  volume={57},
  number={2},
  pages={137--154},
  year={2004},
  publisher={Springer}
}

@inproceedings{zhu2012face,
  title={Face detection, pose estimation, and landmark localization in the wild},
  author={Zhu, Xiangxin and Ramanan, Deva},
  booktitle={IEEE Conference on Computer Vision and Pattern Recognition (CVPR)},
  pages={2879--2886},
  year={2012},
  organization={IEEE}
}

@inproceedings{kruger2000efficient,
  title={Efficient Head Pose Estimation with Gabor Wavelet Networks.},
  author={Kr{\"u}ger, Volker and Sommer, Gerald},
  booktitle={BMVC},
  pages={1--10},
  year={2000},
  organization={Citeseer}
}

@inproceedings{lee2011frequency,
  title={Frequency guided bilateral symmetry Gabor Wavelet Network},
  author={Lee, Seungkyu},
  booktitle={IEEE International Conference on Image Processing (ICIP)},
  pages={281--284},
  year={2011},
  organization={IEEE}
}

@inproceedings{feris2000tracking,
  title={Tracking facial features using Gabor wavelet networks},
  author={Feris, Rogerio S and Cesar, RM},
  booktitle={Brazilian Symposium on Computer Graphics and Image Processing},
  pages={22--27},
  year={2000},
  organization={IEEE}
}

@inproceedings{park2008face,
  title={Face modeling and tracking with Gabor Wavelet Network prior},
  author={Park, Minwoo and Lee, Seungkyu},
  booktitle={International Conference on Pattern Recognition},
  pages={1--4},
  year={2008},
  organization={IEEE}
}

@inproceedings{krueger2000affine,
  title={Affine real-time face tracking using gabor wavelet networks},
  author={Krueger, Volker and Happe, Alexander and Sommer, Gerald},
  booktitle={International Conference on Pattern Recognition},
  volume={1},
  pages={127--130},
  year={2000},
  organization={IEEE}
}

@article{yang2013occluded,
  title={Occluded and low resolution face detection with hierarchical deformable model},
  author={Yang, Xiong and Peng, Gang and Cai, Zhaoquan and Zeng, Kehan},
  journal={Journal of Convergence},
  volume={4},
  number={2},
  pages={11--14},
  year={2013}
}

@article{qahwajia2001detecting,
  title={Detecting Faces in Noisy Images},
  author={Qahwajia, Rami and Green, Roger},
  year={2001},
  publisher={Citeseer}
}

@incollection{zheng2010face,
  title={Face detection in low-resolution color images},
  author={Zheng, Jun and Ram{\'\i}rez, Geovany A and Fuentes, Olac},
  booktitle={Image Analysis and Recognition},
  pages={454--463},
  year={2010},
  publisher={Springer}
}

@inproceedings{hsu2007improved,
  title={An improved face detection method in low-resolution video},
  author={Hsu, Chih-Chung and Chang, Hsuan T and Chang, Ting-Cheng},
  booktitle={International Conference on Intelligent Information Hiding and Multimedia Signal Processing},
  volume={2},
  pages={419--422},
  year={2007},
  organization={IEEE}
}

@incollection{rezaei2014global,
  title={Global Haar-like features: A new extension of classic Haar features for efficient face detection in noisy images},
  author={Rezaei, Mahdi and Nafchi, Hossein Ziaei and Morales, Sandino},
  booktitle={Image and Video Technology},
  pages={302--313},
  year={2014},
  publisher={Springer}
}

@inproceedings{kruppa2003using,
  title={Using Local Context To Improve Face Detection.},
  author={Kruppa, Hannes and Schiele, Bernt},
  booktitle={BMVC},
  pages={1--10},
  year={2003},
  organization={Citeseer}
}

@article{kovesi1999phase,
  title={Phase preserving denoising of images},
  author={Kovesi, Peter},
  journal={signal},
  volume={4},
  number={3},
  pages={1},
  year={1999}
}

@inproceedings{feris2001efficient,
  title={Efficient real-time face tracking in wavelet subspace},
  author={Feris, Rog{\'e}rio S and Cesar Jr, Roberto M and Kr{\"u}ger, Volker},
  booktitle={Recognition, Analysis, and Tracking of Faces and Gestures in Real-Time Systems, 2001. Proceedings. IEEE ICCV Workshop on},
  pages={113--118},
  year={2001},
  organization={IEEE}
}

@inproceedings{viola2001rapid,
  title={Rapid object detection using a boosted cascade of simple features},
  author={Viola, Paul and Jones, Michael},
  booktitle={Proceedings of the 2001 IEEE computer society conference on computer vision and pattern recognition. CVPR 2001},
  volume={1},
  pages={I--I},
  year={2001},
  organization={IEEE}
}

@article{viola2001robust,
  title={Robust real-time object detection},
  author={Viola, Paul and Jones, Michael and others},
  journal={International journal of computer vision},
  volume={4},
  number={34-47},
  pages={4},
  year={2001}
}

@article{freund1997decision,
  title={A decision-theoretic generalization of on-line learning and an application to boosting},
  author={Freund, Yoav and Schapire, Robert E},
  journal={Journal of computer and system sciences},
  volume={55},
  number={1},
  pages={119--139},
  year={1997},
  publisher={Elsevier}
}

@article{lecun1998gradient,
  title={Gradient-based learning applied to document recognition},
  author={LeCun, Yann and Bottou, L{\'e}on and Bengio, Yoshua and Haffner, Patrick},
  journal={Proceedings of the IEEE},
  volume={86},
  number={11},
  pages={2278--2324},
  year={1998},
  publisher={IEEE}
}
\end{rezabib}

\end{document}